\documentclass[letterpaper]{article} % DO NOT CHANGE THIS
\usepackage{aaai2026}  % DO NOT CHANGE THIS
\usepackage{times}  % DO NOT CHANGE THIS
\usepackage{helvet}  % DO NOT CHANGE THIS
\usepackage{courier}  % DO NOT CHANGE THIS
\usepackage[hyphens]{url}  % DO NOT CHANGE THIS
\usepackage{graphicx} % DO NOT CHANGE THIS
\usepackage{natbib}  % DO NOT CHANGE THIS AND DO NOT ADD ANY OPTIONS TO IT
\usepackage{caption} % DO NOT CHANGE THIS AND DO NOT ADD ANY OPTIONS TO IT
\usepackage{algorithm}
\usepackage{algorithmic}
\usepackage{amsmath}
\usepackage{bm}
\usepackage{amssymb}
\usepackage{booktabs} 
\usepackage{newfloat}
\usepackage{listings}
\DeclareCaptionStyle{ruled}{labelfont=normalfont,labelsep=colon,strut=off} % DO NOT CHANGE THIS
\floatstyle{ruled}
\newfloat{listing}{tb}{lst}{}
\floatname{listing}{Listing}
\title{Graph‑Integrated Multimodal Concept Bottleneck Model}
\author{Jiakai Lin, Jinchang Zhang, Guoyu Lu
\thanks{Jiakai Lin, Jinchang Zhang and Guoyu Lu are with the Intelligent Vision and Sensing (IVS) Lab at SUNY Binghamton, USA. 
        {\tt\small guoyulu62@gmail.com}}%
}

\begin{document}

\maketitle

\begin{abstract}
With growing demand for interpretability in deep learning—especially in high‐stakes domains. Concept Bottleneck Models (CBMs) address this by inserting human‐understandable concepts into the prediction pipeline, but they are generally single‐modal and ignore structured concept relationships. To overcome these limitations, we present MoE‐SGT, a reasoning‐driven framework that augments CBMs with a structure‐injecting Graph Transformer and a Mixture‐of‐Experts (MoE) module. 
We construct answer–concept and answer–question graphs for multimodal inputs to explicitly model the structured relationships among concepts. Subsequently, we integrate Graph Transformer to capture multi-level dependencies, addressing the limitations of traditional Concept Bottleneck Models in modeling concept interactions. However, it still encounters bottlenecks in adapting to complex concept patterns. Therefore, we replace the feed-forward layers with a Mixture-of-Experts (MoE) module, enabling the model to have greater capacity in learning diverse concept relationships while dynamically allocating reasoning tasks to different sub-experts, thereby significantly enhancing the model's adaptability to complex concept reasoning. 
MoE-SGT achieves higher accuracy than other concept bottleneck networks on multiple datasets by modeling structured relationships among concepts and utilizing a dynamic expert selection mechanism.
\end{abstract}


\section{Introduction}

Deep learning has achieved remarkable performance in domains such as medical diagnosis, agricultural disease detection, and environmental monitoring. However, the “black-box” nature of many high-performing models poses significant challenges for applications that require safety, reliability, and decision traceability. To build trust in model predictions under high-stakes conditions, it is essential to understand which semantic concepts drive a model’s decisions and to provide mechanisms for human intervention when errors occur.
Concept Bottleneck Models (CBMs) \cite{koh2020concept} address this need by inserting an intermediate concept prediction layer into the network. In CBMs, the model first predicts a set of human-defined, semantically meaningful concepts, and then uses those predicted concepts to make the final task prediction. While the original CBM framework enhances interpretability by making the reasoning process explicit, its reliance on discrete concept slots limits its ability to model relationships among concepts and to fuse multi-modal information in complex settings.
Graph-based architectures have shown great promise in capturing structured dependencies for multimodal reasoning. Graph Convolutional Networks (GCNs) \cite{zhang2019graph} and Graph Attention Networks (GATs) \cite{velivckovic2017graph} introduced convolution and attention mechanisms on graph-structured data, enabling effective modeling of node interactions. More recently, Transformer-style graph models such as Graph Transformer \cite{zhao2021gophormer} and Graphormer \cite{lin2021mesh} embed local subgraph structures and encode edge attributes within self-attention layers, while heterogeneous Graph Transformers extend this paradigm to multiple node and edge types \cite{hu2020heterogeneous, li2019deepgcns}. These graph networks have advanced tasks ranging from visual question answering to commonsense reasoning by explicitly modeling relationships among multimodal elements.

\begin{figure*}[h]
\begin{center}
\includegraphics[width=0.98\textwidth]{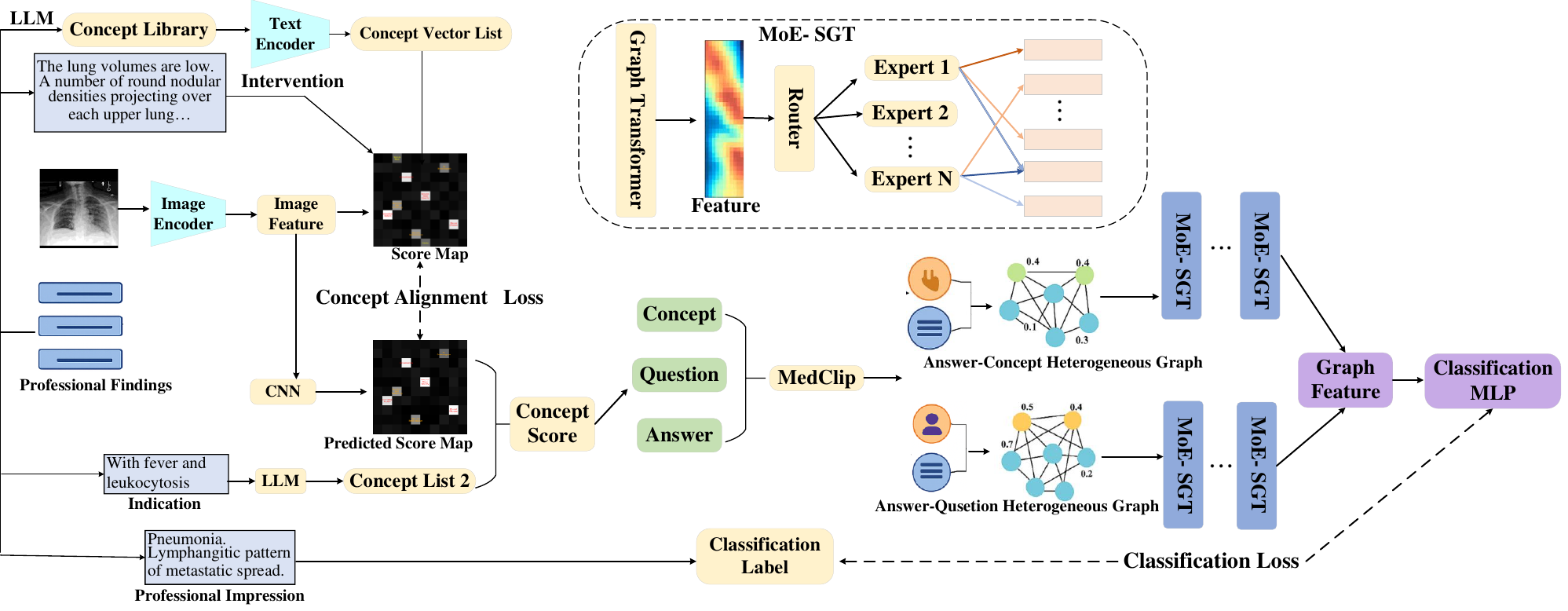}
\end{center}
\caption{Overview of the framework. We extract candidate concepts with an LLM to form the final concept pool. Then we
generate predicted and prior concept scores for each image with optional intervention, and optimize with alignment and
sparsity losses to obtain the multimodal concept bottleneck vector. We generate questions and answers based on the concepts, extract features with MedCLIP, construct answer–concept and answer–question heterogeneous graphs, and then extract graph features through multiple MoE-SGT modules. Finally, these features are fed into an MLP classifier for disease classification.
}
\label{architecture}
\end{figure*}

In this work, we propose MoE-SGT (Mixture-of-Experts Structure-injecting Graph Transformer for Concept Bottleneck Models) illustrated in Figure~\ref{architecture}, a novel framework that unifies concept bottleneck modeling with graph-based multimodal reasoning. MoE-SGT first constructs two heterogeneous graphs—an answer–concept graph and an answer–question graph—where nodes represent visual features, question tokens, and semantic concepts. We then apply a structure-injecting Graph Transformer augmented with dynamic Mixture-of-Experts (MoE) layers, enabling the model to dynamically route computation through specialized expert sub-networks based on input complexity. This design allows MoE-SGT to capture multi-level dependencies across modalities and to adapt its capacity to diverse inputs. Moreover, our framework supports single- or multi-image inputs, automatic or manual concept pool specification, and both ground-truth and pseudo-concept supervision, offering broad applicability from large-scale automated inference to interactive expert interventions.
The main contributions of this paper are as follows:  
1. We introduce the first integration of graph networks into the Concept Bottleneck Model framework, enabling explicit modeling of interactions among semantic concepts to improve both interpretability and reasoning performance.  
2. We design a versatile multimodal architecture that supports automatic concept extraction, ground-truth concept supervision, human-in-the-loop interventions, and multi-image inputs, thereby extending the applicability of CBMs to diverse scientific and industrial reasoning tasks.  
3. We develop a dynamic Mixture-of-Experts mechanism within a Graph Transformer, allowing the model to flexibly adjust its capacity based on input complexity and to facilitate cross-modal reasoning through expert selection.

\section{Related Work}

\subsection{Concept Bottleneck Models}
% Koh et al. \cite{koh2020concept} first proposed Concept Bottleneck Models, which insert discrete concept slots into neural architectures to enforce semantic interpretability. Subsequent work has extended CBMs in multiple directions. Concept Embedding Models replace discrete slots with continuous embedding spaces to improve robustness when annotations are scarce \cite{CEM}. Stochastic CBMs model each concept as a learnable distribution and capture inter-concept covariances, allowing interventions on one concept to influence others \cite{SCBM}. Label-Free CBMs employ CLIP-inferred pseudo-concepts to supervise the bottleneck without requiring ground-truth labels \cite{labelCBM, radford2021learning}. Vision-Language-Guided CBMs use a CLIP-based cross-modal alignment loss to select and jointly represent concepts \cite{vlg-cbm}. Sparse CBMs enforce sparse concept activations via Gumbel-Softmax and align them with text prototypes through contrastive loss \cite{sparseCBM}. Cross-Modal CBMs adopt cross-attention between image patches and text tokens to learn shared concept queries \cite{XBM}. Despite these advances, most CBM variants remain single-modality or do not model structured relationships among concepts, limiting their multi-modal reasoning capabilities.
Koh et al. \cite{koh2020concept} first proposed Concept Bottleneck Models, which insert discrete concept slots into neural architectures to enforce semantic interpretability. Subsequent work has extended CBMs in multiple directions. Concept Embedding Models replace discrete slots with continuous embedding spaces to improve robustness when annotations are scarce \cite{CEM}. Stochastic CBMs model each concept as a learnable distribution and capture inter-concept covariances, allowing interventions on one concept to influence others \cite{SCBM}. Label-Free CBMs employ CLIP-inferred pseudo-concepts to supervise the bottleneck without requiring ground-truth labels \cite{labelCBM, radford2021learning}. Vision-Language-Guided CBMs use a CLIP-based cross-modal alignment loss to select and jointly represent concepts \cite{vlg-cbm}. Sparse CBMs enforce sparse concept activations via Gumbel-Softmax and align them with text prototypes through contrastive loss \cite{sparseCBM}. Cross-Modal CBMs adopt cross-attention between image patches and text tokens to learn shared concept queries \cite{XBM}. More recent variants introduce probabilistic and post-hoc extensions: Probabilistic CBMs model uncertainty in concept predictions within a Bayesian framework \cite{probcbm2024}, while post-hoc CBMs apply bottleneck interventions after training to correct mispredicted concepts \cite{yuksekgonul2022post}. In the medical domain, concept bottleneck with visual concept filtering further refines concept selection by pruning visually irrelevant concepts, leading to more explainable diagnostic models \cite{kim2023concept2}. Despite these advances, most CBM variants remain single-modality or do not fully capture structured relationships among concepts, limiting their multi-modal reasoning capabilities.

\subsection{Graph-Based Multimodal Reasoning}
% Graph-based architectures have demonstrated strong capability in capturing structured dependencies. Graph Convolutional Networks (GCNs) \cite{zhang2019graph} and Graph Attention Networks (GATs) \cite{velivckovic2017graph} introduced graph convolution and attention mechanisms for node classification. Transformer-style graph models, such as Graph Transformer \cite{zhao2021gophormer} and Graphormer \cite{lin2021mesh}, incorporate subgraph structures and pairwise distance encoding into self-attention layers. Heterogeneous Graph Transformers extend these ideas by employing relation-specific attention heads to model multiple node and edge types \cite{hu2020heterogeneous, li2019deepgcns}. Mixture-of-Experts (MoE) modules, exemplified by Switch Transformer \cite{fedus2022switch} and GShard \cite{lepikhin2020gshard}, have been shown to scale capacity through conditional computation. More recently, the Multi-Modal Structure-Embedding Graph Transformer \cite{zhu2023multi} embeds spatial and linguistic priors in heterogeneous answer–vision and answer–question graphs, achieving state-of-the-art results on Visual Commonsense Reasoning benchmarks. However, these graph-based methods have not been combined with the Concept Bottleneck paradigm to facilitate explicit, structured concept interaction in multimodal reasoning—a gap that MoE-SGT addresses.
%Graph-Based Multimodal Reasoning
Graph‐based architectures have proven highly effective in capturing complex, structured dependencies within data \cite{liang2024}. Graph Convolutional Networks (GCNs) \cite{zhang2019graph} and Graph Attention Networks (GATs) \cite{velivckovic2017graph} introduced convolution and attention mechanisms on graph-structured data. Transformer-style graph models such as Graph Transformer \cite{zhao2021gophormer} and Graphormer \cite{lin2021mesh} embed local subgraph structures and encode pairwise distances into self-attention layers, while Heterogeneous Graph Transformers employ relation-specific attention heads for multi-type nodes and edges \cite{hu2020heterogeneous, li2019deepgcns}. Mixture-of-Experts (MoE) modules like Switch Transformer \cite{fedus2022switch} and GShard \cite{lepikhin2020gshard} scale capacity via conditional computation. Recent efforts have extended these ideas to multimodal contexts, including multimodal graphs techniques for integrating image, text, and other modalities \cite{peng2024learning}, and multimodal learning with graphs \cite{ektefaie2023multimodal}. Multimodal Knowledge Graphs (MMKGs) provide rich, structured repositories that fuse entity, visual, and linguistic information \cite{liu2019mmkg}. Applications include graph-based sarcasm explanation, where multi-source semantic graphs guide the generation of explanations for multimodal sarcasm detection \cite{jing2023multi}. More recently, the Multi-Modal Structure-Embedding Graph Transformer \cite{zhu2023multi} embeds spatial and linguistic priors in heterogeneous answer–vision and answer–question graphs, achieving state-of-the-art results on Visual Commonsense Reasoning benchmarks. However, these graph-based methods have not been explored with the Concept Bottleneck paradigm to enable explicit, structured concept interaction in multimodal reasoning—a gap that MoE-SGT addresses.

\section{Language‑Guided Multimodal Concept Generation Layer}
\label{sec:cbl}
%===============================================================

Let $(x,h,y)$ denote an input image, optional hint text, and its ground‐truth label, respectively. Our objective is to map $x$ (optionally enriched by $h$) into a $K$‐dimensional concept space $\mathbf z(x)\in[0,1]^K$, where each dimension is semantically interpretable, amenable to human intervention, and sufficiently informative for downstream classification. This concept generation layer comprises three stages: (a) automatic concept‐pool extraction via a large language model (LLM); (b) intervenable alignment of predicted concept scores with vision–language priors; and (c) fusion of visual and textual concepts into the final bottleneck representation.

\subsection{Concept-Pool Extraction}

Given reports \(\mathcal T=\{s_i\}_{i=1}^N\), we extract candidate phrases via an LLM to form \(\mathcal C_0\).  Each \(c_j\in\mathcal C_0\) is embedded and \(\ell_2\)-normalized by
$ \tilde t_j = g_T(c_j) / \|g_T(c_j)\|_2.$
We deduplicate by merging any \(c_i,c_j\) with \(\cos(\tilde t_i,\tilde t_j)>\tau_c\), yielding \(\mathcal C_1\), where \(\tau_c\) is the redundancy threshold.
For each \(c\in\mathcal C_1\), let
$  s_y(c) = \cos\!\bigl(g_T(c),\,g_T(y)\bigr),\quad y\in\mathcal Y.$
We compute the mean \(\mu(c)\) and standard deviation \(\sigma(c)\) of \(\{s_y(c)\}\):
\begin{equation}
  \mu(c)=\frac{1}{|\mathcal Y|}\sum_{y\in\mathcal Y}s_y(c),\quad
  \sigma(c)=\sqrt{\frac{1}{|\mathcal Y|}\sum_{y\in\mathcal Y}\bigl(s_y(c)-\mu(c)\bigr)^2}.
\end{equation}
With relevance threshold \(\tau_r\), we assign each concept \(c\) a score \(R(c)=\sigma(c)\) if \(\mu(c)\ge\tau_r\), and \(R(c)=0\) otherwise. We then select the top-\(K\) concepts by \(R(c)\) to form the pool \(\mathcal C\). For datasets without reports, we prompt the LLM to generate descriptions for each label name and apply this pipeline unchanged.  

\subsection{Intervenable Concept Alignment}

% For each input image \(x\) (or a set of views \(\{x^{(m)}\}_{m=1}^M\)), we extract \(\ell_2\)-normalized embeddings: 
% $\bm v^{(m)} = {g_I\bigl(x^{(m)}\bigr)}/{\|g_I\bigl(x^{(m)}\bigr)\|_2}.$
% A lightweight head \(h_\theta\) produces per-concept logits \(p_k^{(m)}(x)\) for each view, which we fuse via a small network \(f_\phi\) into a single prediction:
% \begin{equation}
% p_k(x) = f_\phi\bigl(\{\,p_k^{(m)}(x)\}_{m=1}^M\bigr).
% \end{equation}
% Similarly, we compute per-view vision–language priors,
% \begin{equation}
% f_k^{(m)}(x) = \sigma\!\bigl(\bm v^{(m)\top}\,\bm t_k\bigr),
% \end{equation}
% and fuse them to obtain the final prior
% \begin{equation}
% f_k(x) = f_\phi\bigl(\{\,f_k^{(m)}(x)\}_{m=1}^M\bigr).
% \end{equation}

% If a concept label \(a_k=1\) explicitly indicates presence (or \(a_k=0\) absence), or if a hint text clearly implies \(c_k\), we override that fused prior to 1 or 0; otherwise \(f_k(x)\) remains unchanged. Figure~\ref{fig:intervene} illustrates an example of manual intervention during training.
For each input image \(x\) (or a collection of \(M\) views \(\{x^{(m)}\}_{m=1}^M\)), we first extract view-wise embeddings using the visual encoder \(g_I\) and perform \(\ell_2\) normalization:
\[
\bm v^{(m)} = \frac{g_I\bigl(x^{(m)}\bigr)}{\|g_I\bigl(x^{(m)}\bigr)\|_2}.
\]
A lightweight prediction head \(h_\theta\) then produces per-view, per-concept logits \(p_k^{(m)}(x)\). To aggregate multi-view predictions, we introduce a small fusion network \(f_\phi\) that consolidates these logits into a single concept score:
\begin{equation}
p_k(x) = f_\phi\bigl(\{\,p_k^{(m)}(x)\}_{m=1}^M\bigr).
\end{equation}

In parallel, we compute vision–language prior scores for each view based on the concept embedding \(\bm t_k\):
\begin{equation}
f_k^{(m)}(x) = \sigma\!\bigl(\bm v^{(m)\top}\,\bm t_k\bigr),
\end{equation}
and similarly fuse them to obtain the overall prior:
\begin{equation}
f_k(x) = f_\phi\bigl(\{\,f_k^{(m)}(x)\}_{m=1}^M\bigr).
\end{equation}

\begin{figure}[t]
  \centering
\includegraphics[width=0.98\linewidth]{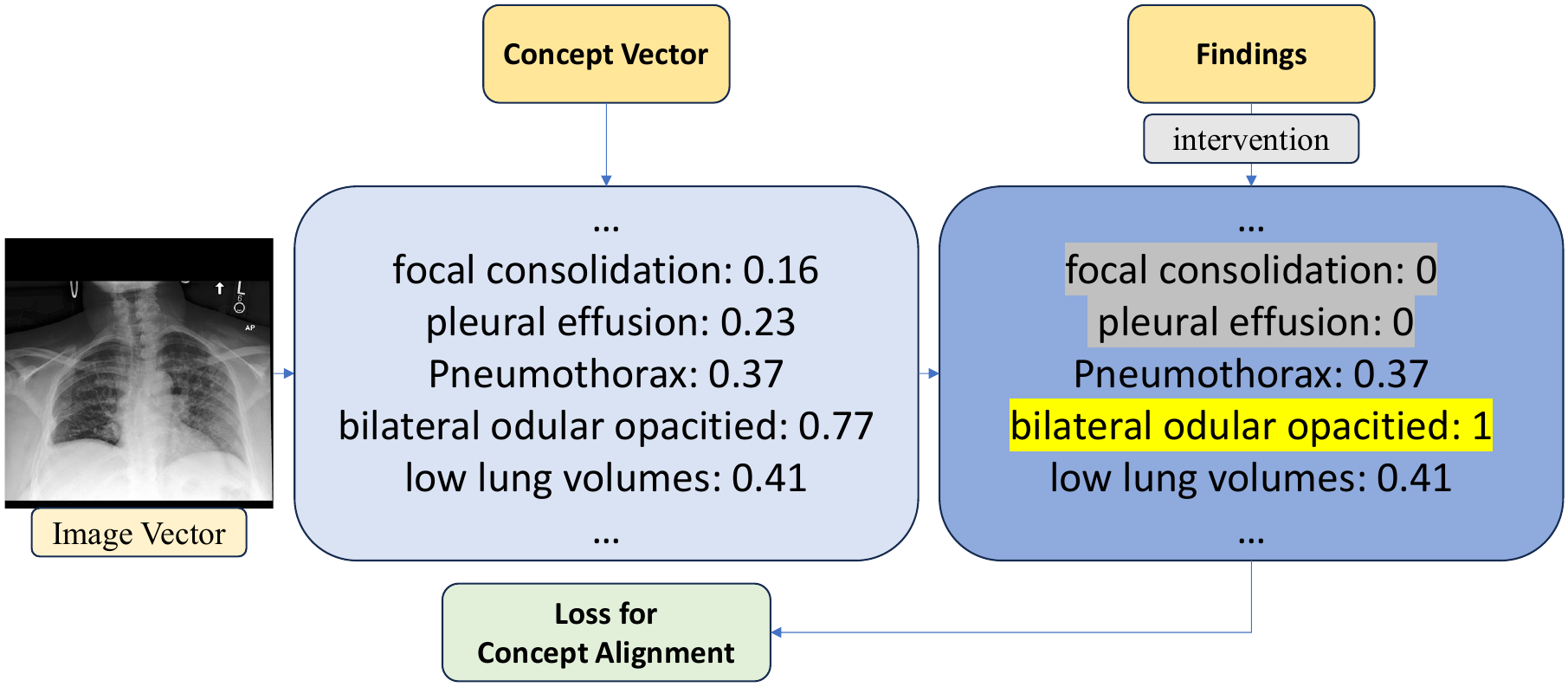}

  \caption{
  % Example of concept‐score overrides %overrides 这个词你再查看一下是不是准确
  %想表达的是“覆盖”
  % during the training intervention phase: after fusing multi‐view priors, explicitly present (or absent) concepts are forced to 1 (or 0).
  Example of concept‐score clamping during the training intervention phase: after fusing multi‐view priors, explicitly present (or absent) concepts are clamped to 1 or 0.
  }
  \label{fig:intervene}
\end{figure}

When a concept label \(a_k=1\) explicitly indicates presence (or \(a_k=0\) indicates absence), or when the hint text \(h\) clearly implies the presence of concept \(c_k\), we override the fused prior \(f_k(x)\) to 1 or 0 for precise human‐in‐the‐loop intervention. Figure~\ref{fig:intervene} provides an example of such manual intervention during training.

% We train the concept head by minimizing the alignment loss:
% \begin{equation}
% \mathcal L_{\mathrm{align}}
% = \frac{1}{K}\sum_{k=1}^K \bigl(p_k(x) - f_k(x)\bigr)^2.
% \end{equation}
% To encourage a parsimonious decision boundary, we regularize the downstream classifier weights \(W_l\) via elastic‐net:
% \begin{equation}
% \mathcal L_{\mathrm{sparse}}
% = \phi\,\|W_l\|_1 + \frac{1-\phi}{2}\,\|W_l\|_F^2,
% \quad
% \phi\in[0,1].
% \end{equation}
% The total bottleneck loss is $\mathcal L_{\mathrm{CBL}}
% = \mathcal L_{\mathrm{align}} + \lambda\,\mathcal L_{\mathrm{sparse}},$
% where \(\lambda\) balances sparsity against alignment.  
We train the concept head by minimizing the alignment loss between predictions and priors:
\begin{equation}
\mathcal L_{\mathrm{align}}
= \frac{1}{K}\sum_{k=1}^K \bigl(p_k(x) - f_k(x)\bigr)^2.
\end{equation}
To encourage a parsimonious decision boundary and improve interpretability, we regularize the downstream classifier weights \(W_l\) with an elastic‐net penalty:
\begin{equation}
\mathcal L_{\mathrm{sparse}}
= \phi\,\|W_l\|_1 + \frac{1-\phi}{2}\,\|W_l\|_F^2,
\quad
\phi\in[0,1],
\end{equation}
where \(\phi\) balances \(L_1\) and \(L_2\) regularization.  

The total bottleneck loss is then defined as
\begin{equation}
\mathcal L_{\mathrm{CBL}}
= \mathcal L_{\mathrm{align}} + \lambda\,\mathcal L_{\mathrm{sparse}},
\end{equation}
where \(\lambda\) trades off alignment accuracy against model sparsity to yield a concept representation that is both interpretable and robust.  

% \subsection{Multimodal Concept Fusion}
% \label{subsec:fusion}

% If a hint text $h$ accompanies $x$,
% its embedding is compared against the prototypes.
% When the maximum cosine exceeds
% $\tau_h$,
% the corresponding prediction is clamped to one,
% $p_{k^{\star}}(x)\!\leftarrow\!1$.
% The multimodal concept vector is therefore
% \begin{equation}
% \label{eq:bottleneck}
% \mathbf z(x)=
% \bigl[p_{1}(x),\dots,p_{K}(x)\bigr]^{\!\top}\in[0,1]^{K},
% \end{equation}
% which jointly encodes
% \emph{visual evidence},
% \emph{language priors}, and
% \emph{explicit human intervention}.
% This vector is the sole input to the graph‑integrated classifier introduced in Section~\ref{sec:ctg}.
\subsection{Multimodal Concept Fusion}
\label{subsec:fusion}

In certain tasks, each input image \(x\) may be accompanied by auxiliary hint text \(h\), which can provide complementary semantic cues. We encode \(h\) via the text encoder \(g_T\) and compute its cosine similarity with each concept prototype \(\{\bm t_k\}_{k=1}^K\). If there exists an index \(k^\star\) such that  
\[
\cos\bigl(g_T(h),\,\bm t_{k^\star}\bigr) > \tau_h,
\]  
we interpret this as a strong indication of concept \(c_{k^\star}\) and explicitly clamp its predicted score to unity:
\[
p_{k^\star}(x) \leftarrow 1.
\]  
This intervention mechanism ensures that human‐provided hints directly influence the model’s reasoning.

After applying any text‐based interventions, we assemble the final multimodal concept vector as  
\begin{equation}
\label{eq:bottleneck}
\mathbf z(x)
=
\bigl[p_{1}(x),\,p_{2}(x),\,\dots,\,p_{K}(x)\bigr]^{\!\top}
\in[0,1]^{K},
\end{equation}
where each entry \(p_k(x)\) integrates three sources of information: (1) visual evidence from the image encoder, (2) learned language priors via concept alignment, and (3) explicit human intervention based on \(h\). The resulting vector \(\mathbf z(x)\) serves as the sole input to the graph‐integrated classifier detailed in the section of Conceptual-Textual Graph, enabling structured, high‐order reasoning over concepts.  

\section{MoE-Enhanced Graph Transformer for Reasoning-driven Classification}

\subsection{Conceptual-Textual Graph}
\label{sec:ctg}
To construct the initial graph structure, the input \emph{concepts} and \emph{texts} are first pre-processed at the \emph{concept level} and the \emph{word level}, respectively, to build the graphs. 
Given \(N_c\) concepts
\(\mathcal{C} = \{c_i\}_{i=1}^{N_c}\),
we use a pretrained concept encoder (such as knowledge-graph embeddings or concept vectors from large language models) to extract their \emph{concept features}
\(\mathcal{F} = \{f_i\}_{i=1}^{N_c}, \ f_i \in \mathbb{R}^d,\)
where \(d\) is the feature dimension. To enhance the discriminative power of the concept representations, additional metadata such as hierarchical categories and attribute tags are embedded into the feature vectors.

For a question containing \(N_q^u\) words
\(\mathcal{U}_q = \{u_i^q\}_{i=1}^{N_q^u}\)
and an answer containing \(N_a^u\) words
\(\mathcal{U}_a = \{u_i^a\}_{i=1}^{N_a^u},\)
their linguistic features are represented as
\(\mathcal{T}_q = \{t_i^q\}_{i=1}^{N_q^u}, \ t_i^q \in \mathbb{R}^d,\)
and
\(\mathcal{T}_a = \{t_i^a\}_{i=1}^{N_a^u}, \ t_i^a \in \mathbb{R}^d,\)
respectively. These features are extracted using a MedCLIP\cite{wang2022medclip} text encoder.
Since our concept classification task is answer-oriented, it is crucial to model the correlations between answers and concepts as well as between answers and questions. 
Motivated by this, we construct two heterogeneous graphs: the answer–concept heterogeneous graph
$G_{ac} = (V_{ac}, E_{ac})$
and the answer–question heterogeneous graph
$ G_{aq} = (V_{aq}, E_{aq}).$
For \(G_{ac}\), the node set \(V_{ac}\) consists of concept regions and words from the answer, with node features initialized from the concept feature set
\(\mathcal{S} = \{s_i\}_{i=1}^{N_o}\)
and the answer features
\(\mathcal{T}_a = \{t_i^a\}_{i=1}^{N_a^u}\).
Edges are constructed in three categories: edges between concept nodes, edges between answer-word nodes, and cross-modal edges between concept and answer nodes.
For the \(N_o\) concept nodes, the edge weights are computed using the Euclidean distance between regions, with two regions located at \((X_i, Y_i)\) and \((X_j, Y_j)\). The edge weight is defined as:
\begin{equation}
D_{ij}^v = l_v \cdot \bigl[(X_i - X_j)^2 + (Y_i - Y_j)^2\bigr],
\quad
E_{ij}^v = \Psi_v(D_{ij}^v),
\end{equation}
% \noindent where \(X, Y\) are normalized to \([0,1]\).
\noindent where, \(l_v\) is a scaling factor to adjust the concept distance to a suitable range, and \(\Psi_v\) is the embedding function for the concept edge weights.
For answer–answer edges, word-order information is incorporated into the edge weight construction. 
% For example, given a sentence with three words, the edge weight matrix is:
% \begin{equation}
% E_a^{3 \times 3} = \Psi_a
% \begin{bmatrix}
% 0 & 1 & 2 \\
% 1 & 0 & 1 \\
% 2 & 1 & 0
% \end{bmatrix},
% \end{equation}
% \noindent where \(\Psi_a\) is the embedding function for answer edge weights. 
Cross-modal edges between concept nodes and answer nodes are uniformly assigned a value of 1 and embedded via \(\Psi_a\), allowing strong cross-modal information flow. The embedded results of these three types of edges form the structural embedding matrix
$E_{ac} \in \mathbb{R}^{(N_o + N_a^u) \times (N_o + N_a^u)}.$
Similarly, for the answer–question graph \(G_{aq}\), the nodes consist of \(N_q^u\) question-word nodes and \(N_a^u\) answer-word nodes. The construction of edge weights follows the same method as above for \(E_a\). The edges between answer words and question words are also assigned a constant value of 1 and embedded 
, yielding the structural embedding matrix
$E_{aq} \in \mathbb{R}^{(N_q^u + N_a^u) \times (N_q^u + N_a^u)}.$
In summary, these heterogeneous graphs use fine-grained concept regions and linguistic tokens as nodes, capturing intra-modal and cross-modal relationships through structural embeddings between concepts and language data. This provides a foundation for subsequent classification tasks.
\subsection{MoE-based Graph Transformer}

To replace traditional graph convolutional networks (GCNs), we propose a Mixture-of-Experts-based Structure-Injecting Graph Transformer (MoE-SGT). This model aims to effectively learn node representations and graph structure evolution from heterogeneous graphs during the contextualization and reasoning processes.

\begin{figure*}[t]
  \centering
  \includegraphics[width=0.97\textwidth]{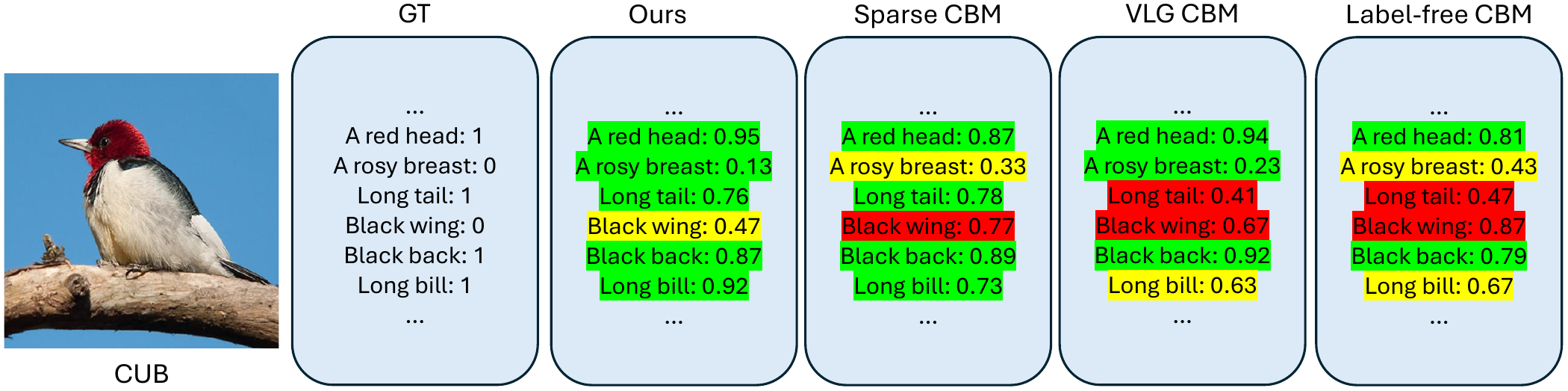}
  \caption{Example of normalized concept score outputs on the CUB dataset for our model, Sparse‐CBM\cite{sparseCBM}, VLG‐CBM\cite{vlg-cbm}, and Label‐Free CBM\cite{labelCBM}. Concepts whose error relative to the ground truth is within 0.3 are highlighted in \textcolor{green}{green (accurate)}, those with error between 0.3 and 0.5 in \textcolor{yellow}{yellow (inaccurate)}, and those with error above 0.5 in \textcolor{red}{red (wrong)}.}
  \label{fig:concept_scores}
\end{figure*}
\subsubsection{Contextualization}

MoE-SGT first performs contextualization between answer words and concept regions, as well as between answer words and question words, using the answer-concept heterogeneous graph and the answer-question heterogeneous graph. For the answer-concept heterogeneous graph \(G_{ac}\), the node representations \(V_{ac}\) and the structure embedding prior \(E_{ac}\) are fed into the SGT layer. To model intra-modal and cross-modal correlations, three fully connected layers are applied to the original node features to generate the query matrix \(M_{ac}^{query} \in \mathbb{R}^{(N_o^u + N_a^u) \times d}\), key matrix \(M_{ac}^{key} \in \mathbb{R}^{(N_o^u + N_a^u) \times d}\), and value matrix \(M_{ac}^{value} \in \mathbb{R}^{(N_o^u + N_a^u) \times d}\).

In a conventional Transformer, the semantic correlation matrix is computed solely based on the semantic similarity between \(M_{query}\) and \(M_{key}\). Although the input features include positional encodings indicating the order of words, this approach is not suitable for cross-modal data. Therefore, in the first SGT layer, the multi-modal structure embedding representation \(E_{ac}\) is introduced as the structural prior \(E^{st}\) into the semantic correlation matrix to jointly generate the node aggregation matrix:
\begin{equation}
E^{st} = E_{ac},
E^{evo}_{ac} = \text{softmax}\Big(E^{st} + \frac{M_{ac}^{query} \cdot (M_{ac}^{key})^\top}{\sqrt{d}}\Big),
\end{equation}
For subsequent SGT layers, the structural prior \(E^{st}\) is scaled by a factor \(l_{sgt}\) for normalization and embedded via the function \(\Psi_{sgt}\):
$E^{st} = \Psi_{sgt}\big(l_{sgt} \cdot E^{evo}_{ac}\big). $
Node features are updated through the node aggregation matrix 
$V^{update}_{ac} = E^{evo}_{ac} \cdot M_{ac}^{value}.$
In this process, the multi-head attention mechanism from the original Transformer is employed to enable diverse embedding learning. The updated node features are then refined through an MoE module to generate the final evolutionary node representations \(V^{evo}_{ac}\) in Section \ref{MoE}.
The evolutionary answer-concept heterogeneous graph representation output by the SGT layer is denoted as:
$G^{evo}_{ac} = \{V^{evo}_{ac}, E^{evo}_{ac}\}.$
\(G^{evo}_{ac}\) is fed into the next SGT layer, and after multiple SGT layers, the final graph representation is obtained:
$G^{final}_{ac} = \{V^{final}_{ac}, E^{final}_{ac}\}. $
Similarly, the original answer-question heterogeneous graph \(G_{aq}\) is processed using the same approach to produce the final graph representation:
$G^{final}_{aq} = \{V^{final}_{aq}, E^{final}_{aq}\}.$

% \subparagraph{2) Reasoning: }
\subsubsection{Reasoning}
After contextualizing answer words with concept regions and question words, the concept-guided answer word representation \(V^{final}_{ac\_a} \in \mathbb{R}^{N_a^u \times d}\) and the question-guided answer word representation \(V^{final}_{aq\_a} \in \mathbb{R}^{N_a^u \times d}\) are selected. These are concatenated with the original answer word representation \(T_a\) to obtain the reasoning feature:$ V^{reason}_{a} = \text{concat}\big(T_a, V^{final}_{ac\_a}, V^{final}_{aq\_a}\big). $

The evolutionary reasoning graph \(G^{reason} = \{V^{reason}_{a}, E^{reason}_{a}\}\) is then constructed based on \(V^{reason}_{a}\) to model the correlation between reasoning answer words, similar to MSE. Subsequently, sequential SGT layers are applied to the reasoning graph for evolutionary learning, resulting in the final node representation \(V^{cls} \in \mathbb{R}^{N_a^u \times 3d}\) and the final graph representation \(E^{cls} \in \mathbb{R}^{N_a^u \times N_a^u}\):
\begin{equation}
\label{12}
\{V^{cls}, E^{cls}\} = \Phi^{(n)}\big(\{V^{reason}_{a}, E^{reason}_{a}\}\big), 
\end{equation}
where \(\Phi^{(n)}(\cdot)\) denotes \(n\) layers of SGT operations.
% Since the reasoning graph is relatively smaller than the answer-concept or answer-question heterogeneous graphs, the number of SGT layers for reasoning is set to 2 to avoid overfitting. Thus, Eq.\ref{12} can be further written as:
% \begin{equation}
% \{V^{cls}, E^{cls}\} = \Phi\Big(\Phi\big(\{V^{reason}_{a}, E^{reason}_{a}\}\big)\Big).
% \end{equation}
In this task, the model needs to process complex multi-modal inputs (including medical concepts, questions, and answers) and fully capture the multi-level relationships among concepts as well as cross-modal interactions. The Graph Transformer struggles to simultaneously handle diverse concept types and their associated information. To address this issue, we introduce a Mixture-of-Experts (MoE) module with K experts into the Transformer module, replacing the traditional FFN block. The MoE module employs a dynamic gating mechanism to allocate input data to multiple sub-expert networks. Each expert is responsible for extracting features and performing reasoning within its assigned data subspace, thereby significantly enhancing the model's representational capacity and diversity. The output of the MoE module can be expressed as:
\begin{equation}   
\text{MoE}(V_{ac}^{update}) = \sum_{k=1}^{K} G_k(V_{ac}^{update}) \cdot E_k(V_{ac}^{update}),
\label{MoE}
\end{equation}
\noindent where \(E_k(\cdot)\) denotes the \(k\)-th expert sub-network (typically with a similar structure to the FFN); \(G_k(\cdot)\) represents the gating network, which outputs the weight of the \(k\)-th expert for the given input sample; \(K\) is the total number of experts.
The gating network \(G_k(\cdot)\) typically employs softmax normalization to ensure the sum of all expert weights equals 1:
\begin{equation}   
G_k(V_{ac}^{\text{update}}) = \frac{\exp(g_k(V_{ac}^{\text{update}}))}{\sum_{j=1}^{K} \exp(g_j(V_{ac}^{\text{update}}))}, 
\end{equation}
where \(g_k(\cdot)\) is the scoring function for the \(k\)-th expert, which can be implemented using a lightweight MLP or dot-product attention.
By introducing the MoE module, we not only largely enhance the parameter capacity and non-linear representational capability of the model, but also enable stronger expert selection abilities during cross-modal reasoning.

\section{Experiments}
\subsection{Datasets and Settings}
%下面对于数据集的描述可以去掉，或者简写。refer到具体的paper去看细节。
% We evaluate on both multi‐label and single‐label benchmarks. For multi‐label classification, we use MIMIC‐CXR \cite{mimic} and CheXpert \cite{chexpert}. For single‐label tasks, we employ CUB-200 \cite{cub} and CIFAR-100 \cite{cifar100}. In MIMIC‐CXR and CUB-200—which provide explicit concept ground truth—we first extract an LLM‐based concept pool and then manually force any annotated concept to its maximum score during training. In both multi‐label datasets, all “uncertain” labels are mapped to negative. We set the number of concepts to 1,024 for MIMIC‐CXR and CheXpert, 312 for CUB-200, and 824 for CIFAR-100. For MIMIC‐CXR and CheXpert, we use MedCLIP\cite{wang2022medclip} for higher modality alignment; for CUB-200 and CIFAR-100, we use standard CLIP\cite{radford2021learning} for both text and image encoding. We apply Deepseek-r1 (32b) model\cite{guo2025deepseek} as the LLM for all datasets. We set \(\tau_r\)=0.85, \(\tau_c\) = 0.1 and the number of expert to 8 for our model.
We evaluate our framework on both multi‐label and single‐label benchmarks. For multi‐label classification, we choose MIMIC‐CXR \cite{mimic} and CheXpert \cite{chexpert}. For single‐label tasks, we employ CUB‐200 \cite{cub}, ImageNet \cite{imagenet}, CIFAR-10 and CIFAR-100  \cite{cifar100}. Among these, only CUB‐200 provides an official concept pool. In the concept‐annotated datasets (MIMIC‐CXR and CUB-200), we first extract a concept pool via an LLM and then manually enforce any ground‐truth concept to its maximum score during training. And all “uncertain” labels are treated as negative. We set the number of concepts to 1,024 for MIMIC-CXR and CheXpert, 312 for CUB-200, 824 for CIFAR-100, 4,086 for ImageNet, and 120 for CIFAR-10. For MIMIC-CXR and CheXpert, we use MedCLIP \cite{wang2022medclip} to enhance cross‐modal alignment; for CUB-200, CIFAR-100, ImageNet, and CIFAR-10, we use standard CLIP \cite{radford2021learning} for both text and image encoding. Across all datasets, we employ the Deepseek-r1 (32b) model \cite{guo2025deepseek} as our LLM. We set \(\tau_r = 0.85\), \(\tau_c = 0.1\), and use 8 experts in the MoE module. 

We compare our approach against six state-of-the-art or representative open-source CBM variants—Sparse-CBM \cite{sparseCBM}, Label-Free CBM \cite{labelCBM}, VLG-CBM \cite{vlg-cbm}, SCBM \cite{SCBM}, CEM \cite{CEM}, and the original CBM \cite{koh2020concept}. To ensure a fair comparison, all models are trained and evaluated on the official training/validation/test splits provided by each dataset. For the datasets without predefined concept sets, models that automatically generate a concept pool (Sparse‐CBM \cite{sparseCBM}, Label‐Free CBM \cite{labelCBM}, VLG‐CBM \cite{vlg-cbm}, SCBM \cite{SCBM}) retain their original concept‐mining procedures, while CEM \cite{CEM} and the original CBM \cite{koh2020concept} use the same pre‐generated pool as our method. 

% For the multi-label benchmarks MIMIC-CXR and CheXpert, we adapt each model’s final activation layer by replacing the single-class softmax with an independent sigmoid head of dimension equal to the number of pathologies, and we substitute the corresponding multi-label binary cross-entropy loss for the original single-label cross-entropy. Since these two datasets provide frontal and lateral views for each study (whereas most compared models expect a single image), we run each model twice per study, once on the frontal view and once on the lateral view, and take the element-wise maximum of the two output probability vectors as the final prediction. For the  datasets without predefined concept sets, models that automatically generate a concept pool (Sparse-CBM \cite{sparseCBM}, Label-Free CBM \cite{labelCBM}, VLG-CBM \cite{vlg-cbm}, and SCBM \cite{SCBM}) retain their original concept-mining procedures; for CEM \cite{CEM} and CBM \cite{koh2020concept}, which require an explicit concept pool, we use the same pre-generated pool as in our method. Performance on MIMIC-CXR and CheXpert is measured by ROC-AUC and F1-score, whereas CUB-200, ImageNet, CIFAR-10 and CIFAR-100 are evaluated by Top-1 accuracy. Each model is trained for 50 epochs and then evaluated on the same held-out test set; the results are reported in Table  \ref{tab:comparison}.

\subsection{Single‐Label Classification Results}
 Performance is measured by Top‐1 accuracy and results are summarized in Table~\ref{tab:comparison}. Figure \ref{fig:concept_scores} shows one of the prediction from CUB dataset. 
It is evident that our model achieves the best Top‐1 accuracy on three of the four datasets. On CUB‐200—where a fixed, official concept pool is provided—our method attains 79.76\% accuracy, only 0.51\% below the best baseline (Sparse‐CBM’s 80.27\%). This small gap likely reflects that CUB‐200’s fixed concept set limits the benefits of our automatic concept extraction and concept‐graph weighting. Nevertheless, our model still substantially outperforms the remaining variants. On ImageNet, CIFAR‐10, and CIFAR‐100—where concept pools must be mined or supplied—our approach surpasses all baselines, demonstrating the effectiveness of our unified multimodal architecture and dynamic MoE routing in diverse single‐label tasks.
\begin{table}[t]
\centering
\resizebox{\linewidth}{!}{%
\begin{tabular}{lcccc}
\toprule
\textbf{Model} & \shortstack{\textbf{CUB-200}\\\textbf{(Accuracy \%)}} & 
\shortstack{\textbf{ImageNet}\\\textbf{(Accuracy \%)}} & 
\shortstack{\textbf{CIFAR-10}\\\textbf{(Accuracy \%)}} & 
\shortstack{\textbf{CIFAR-100}\\\textbf{(Accuracy \%)}} \\
\midrule
Ours               & 79.76 & \textbf{73.41} & \textbf{92.41}  & \textbf{76.89} \\
Sparse-CBM\cite{sparseCBM}         & \textbf{80.27} & 71.32 & 91.08 & 74.23 \\
Label-free CBM\cite{labelCBM}     & 72.13 & 71.62 & 86.04 & 65.07 \\
VLG-CBM\cite{vlg-cbm}            & 74.21 & 67.19 & 81.72 & 63.17 \\
SCBM\cite{SCBM}               & 68.31 & 67.22 & 70.67 & 59.31 \\
CEM\cite{CEM}                & 69.29 & 68.63 & 72.35 & 60.79 \\
CBM\cite{koh2020concept}                & 67.35 & 65.84 & 69.38 & 58.11 \\
\bottomrule
\end{tabular}%
}
\caption{Top-1 accuracy (\%) of our model and six benchmarks—Sparse CBM \cite{sparseCBM}, Label-Free CBM \cite{labelCBM}, VLG-CBM \cite{vlg-cbm}, SCBM \cite{SCBM}, CEM \cite{CEM} and CBM \cite{koh2020concept} on four single‐label benchmarks. The best result for each dataset is shown in bold.}
\label{tab:comparison}
\end{table}
\subsection{Multi‐Label Classification Results}

To adapt each model to the multi‐label setting, we replace their final single‐class softmax with a sigmoid head of dimension equal to the number of pathologies, and adopt a multi‐label binary cross‐entropy loss. Since both datasets include frontal and lateral views per study, we run each model on both views and take the element‐wise maximum of the two output probability vectors as the final prediction, treating all “uncertain” labels as negative. Models are trained for 50 epochs on the official training splits and evaluated by ROC‐AUC and F1‐score; results appear in Table~\ref{tab:multi}.

Our method obtains an ROC‐AUC of 0.76 and an F1 score of 0.80 on MIMIC‐CXR, outperforming the strongest baseline, Sparse‐CBM (0.72/0.73), by margins of 0.04 and 0.07. Similarly, on CheXpert we achieve 0.83 AUC and 0.86 F1, surpassing Sparse‐CBM’s 0.80 AUC and 0.81 F1 by 0.03 and 0.05, respectively. Notably, the improvement in F1 is consistently larger than that in AUC, indicating that our model not only ranks positive cases more accurately but also strikes a better balance between precision and recall under label ambiguity. These results highlight the effectiveness of combining explicit concept reasoning with adaptive expert routing: the graph‐based bottleneck captures structured co‐occurrence patterns among pathologies, while the dynamic MoE mechanism allocates model capacity where it is most needed, leading to more robust predictions across both datasets.

% \begin{table}[t]
% \centering
% \resizebox{\linewidth}{!}{%
% \begin{tabular}{lcccc}
% \toprule
% \textbf{Model} & \shortstack{\textbf{MIMIC-CXR}\\\textbf{(AUC)}} & \shortstack{\textbf{CheXpert}\\\textbf{(AUC)}} & \shortstack{\textbf{CUB-200}\\\textbf{(Accuracy \%)}} & \shortstack{\textbf{CIFAR-100}\\\textbf{(Accuracy \%)}} \\
% \midrule
% Ours               & \textbf{0.76} & \textbf{0.83} & 79.76 & \textbf{76.89} \\
% Sparse-CBM\cite{sparseCBM}         & 0.72          & 0.80          & \textbf{80.27}          & 74.23          \\
% Label-free CBM\cite{labelCBM}     & 0.70          & 0.76          & 72.13          & 65.07          \\
% VLG-CBM\cite{vlg-cbm}            & 0.71          & 0.78          & 74.21          & 63.17          \\
% SCBM\cite{SCBM}               & 0.67          & 0.71          & 68.31          & 59.31          \\
% CEM\cite{CEM}                & 0.68          & 0.71          & 69.29          & 60.79          \\
% CBM\cite{koh2020concept}                & 0.66          & 0.69          & 67.35          & 58.11          \\
% \bottomrule
% \end{tabular}%
% }
% \caption{Evaluation results of our model and six benchmarks—Sparse, CBM \cite{sparseCBM}, Label-Free CBM \cite{labelCBM}, VLG-CBM \cite{vlg-cbm}, SCBM \cite{SCBM}, CEM \cite{CEM} and CBM \cite{koh2020concept}, on four datasets. The best result for each dataset is shown in bold.}
% \label{tab:comparison}
% \end{table}

\begin{table}[t]
\centering
\resizebox{\linewidth}{!}{%
\begin{tabular}{lcccc}
\toprule
\textbf{Model} & \shortstack{\textbf{MIMIC-CXR}\\\textbf{(AUC)}} & \shortstack{\textbf{MIMIC-CXR}\\\textbf{(F1)}} & \shortstack{\textbf{CheXpert}\\\textbf{(AUC)}} & \shortstack{\textbf{CheXpert}\\\textbf{(F1)}}\\
\midrule
Ours               & \textbf{0.76} & \textbf{0.80}  & \textbf{0.83} & \textbf{0.86} \\
Sparse-CBM\cite{sparseCBM}         & 0.72   & 0.73 & 0.80          &  0.81 \\ 
Label-free CBM\cite{labelCBM}     & 0.70          & 0.71 & 0.76          & 0.76 \\
VLG-CBM\cite{vlg-cbm}            & 0.71          & 0.73 & 0.78          & 0.80 \\
SCBM\cite{SCBM}               & 0.67          & 0.68 & 0.71          &  0.74\\
CEM\cite{CEM}                & 0.68          & 0.70 & 0.71          &  0.75\\
CBM\cite{koh2020concept}                & 0.66          & 0.67 & 0.69          & 0.72 \\
\bottomrule
\end{tabular}%
}
\caption{ROC-AUC and F1 scores of our model and six benchmarks—Sparse CBM \cite{sparseCBM}, Label-Free CBM \cite{labelCBM}, VLG-CBM \cite{vlg-cbm}, SCBM \cite{SCBM}, CEM \cite{CEM} and CBM \cite{koh2020concept} on  multi‐label benchmarks. The best result for each metric is shown in bold.}
\label{tab:multi}
\end{table}

 % It is evident that our model achieves the best performance on three of the four datasets. On the two multi‐label benchmarks, our method surpasses all baselines by a substantial margin. In the CUB-200 fine-grained bird classification task, our model’s Top-1 accuracy is only 0.51\% lower than the best baseline (Sparse-CBM \cite{sparseCBM}). This small gap likely arises because CUB-200 provides a fixed concept pool, preventing our approach from fully exploiting its strength in automatically generating superior prior concepts and learning concept-graph weights. Nonetheless, our model still markedly outperforms the remaining models.

\begin{table}[ht]
\begin{center}
\resizebox{\linewidth}{!}{
\begin{tabular}{llllllll}
\hline 
\multicolumn{1}{c}{QA} &\multicolumn{1}{c}{Graph Transformer}  &\multicolumn{1}{c}{MoE}  &\multicolumn{1}{c}{\bf MIMIC-CXR (AUC)} &\multicolumn{1}{c}{\bf CUB-200 (Accuracy \% )} 
\\  \hline   
\     &   &  & 0.69  &   70.31
\\  \hline   
\   \checkmark  &   &  & 0.72  &   76.67
\\  \hline 
 \checkmark&\checkmark  & &0.75 &78.74 
\\  \hline  
 \checkmark &\checkmark & \checkmark       &   0.76  &   79.76
\\  \hline  

\end{tabular}}
\end{center}
\caption{We conduct an ablation study on the MIMIC-CXR dataset and CUB-200 to evaluate the contribution of each module in the proposed. }
\label{Ablation_1}
\end{table}

\subsection{Ablation Study}
\subsubsection{Module Contribution Ablation}
As shown in Table~\ref{Ablation_1}, our base model without the QA modality, Graph Transformer or MoE module achieves an AUC of 0.69 on MIMIC-CXR and a Top-1 accuracy of 70.31\% on CUB-200 (Row 1). Adding the QA modality (Row 2) yields the largest single improvement, improving AUC by 0.03 to 0.72 and accuracy by 6.36 percentage points to 76.67\%, which underscores the critical role of question–answer information in modeling high-level dependencies and enabling cross-modal reasoning. Replacing the simple GCN with our structure-injecting Graph Transformer (Row 3) further boosts AUC by 0.03 to 0.75 and accuracy by 2.07 percentage points to 78.74\%, confirming that explicitly capturing structured inter-concept relationships in heterogeneous graphs significantly enhances representational power. Finally, incorporating the dynamic MoE module (Row 4) raises AUC by 0.01 to 0.76 and accuracy by 1.02 percentage points to 79.76\%, demonstrating that expert routing is essential for adapting model capacity to diverse feature spaces and input distributions. Overall, the QA modality contributes the largest gain, the Graph Transformer provides substantial structured-reasoning improvements, and the MoE module delivers consistent fine-tuning benefits.  

\subsubsection{MoE Expert Count Ablation on Single‐Label Tasks}
\begin{figure}[t]
  \centering
\includegraphics[width=0.98\linewidth]{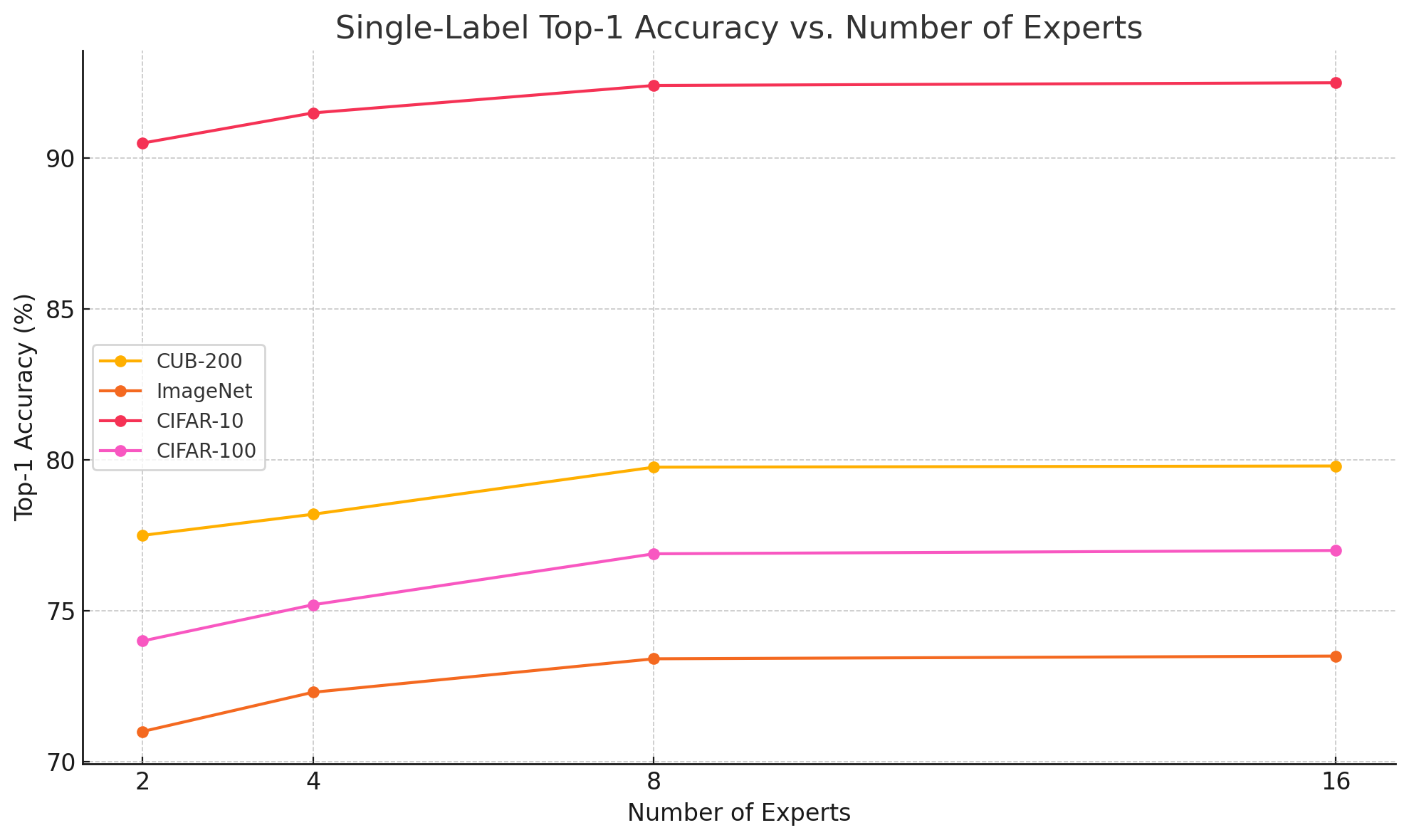}
  \caption{Single-label Top-1 accuracy regarding the number of MoE experts on CUB-200, ImageNet, CIFAR-10, and CIFAR-100.}
  \label{fig:top1_experts}
\end{figure}

To determine the optimal number of experts in our MoE module, we vary the expert count from 2 to 16 while keeping QA modality and Graph Transformer fixed. Figure~\ref{fig:top1_experts} plots Top-1 accuracy on four single-label benchmarks (CUB-200, ImageNet, CIFAR-10, CIFAR-100) against the number of experts. We observe that increasing from 2 to 8 experts yields the largest gains: CUB-200 rises from 77.50\% to 79.76\%, ImageNet from 71.00\% to 73.41\%, CIFAR-10 from 90.50\% to 92.41\%, and CIFAR-100 from 74.00\% to 76.89\%. Beyond 8 experts, improvements plateau (e.g.\ CUB-200 only to 79.80\% at 16 experts), indicating diminishing returns. Thus, 8 experts strike the best balance between capacity and efficiency.

\subsubsection{MoE Expert Count Ablation on Multi‐Label Tasks}

\begin{figure}[t]
  \centering
\includegraphics[width=0.98\linewidth]{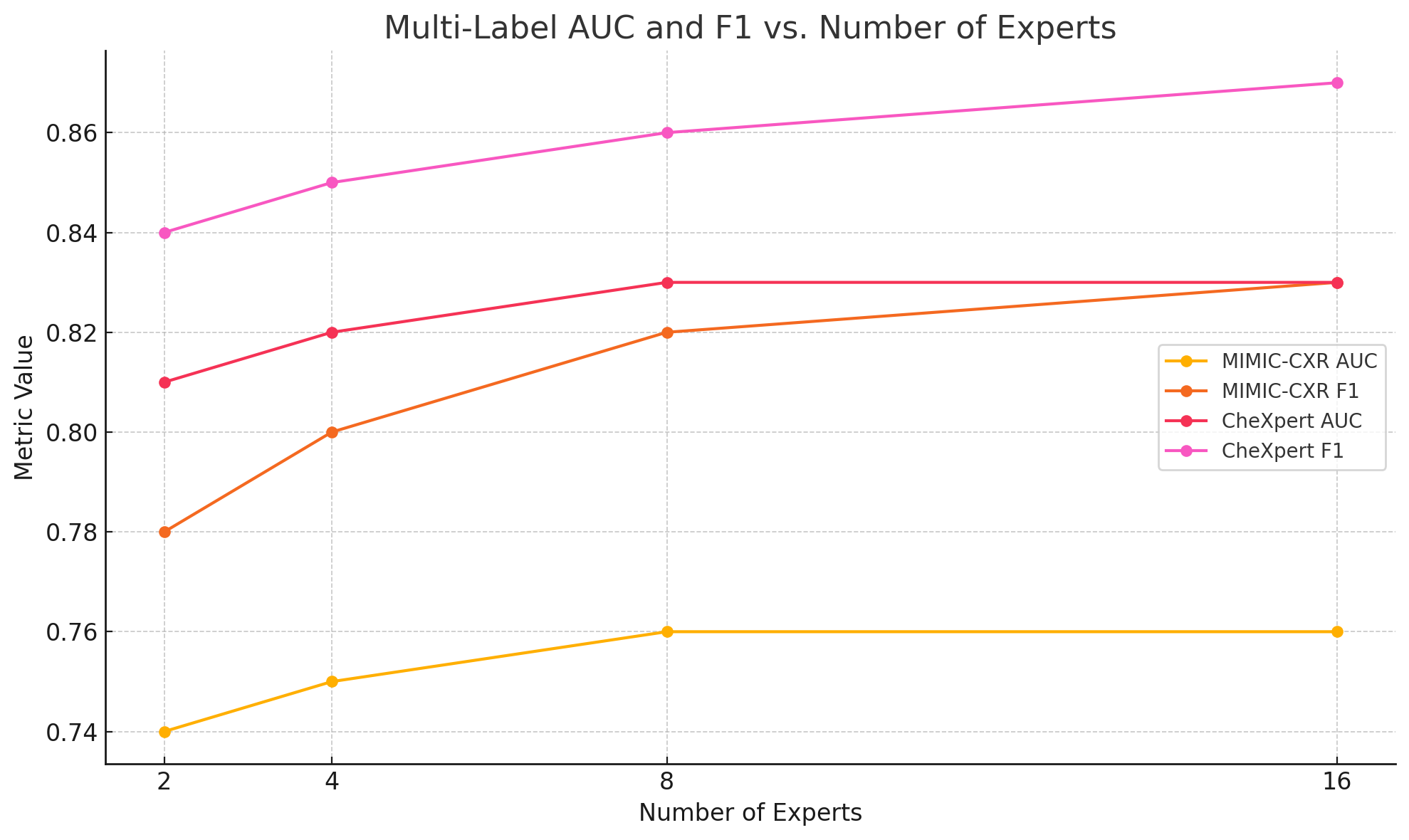}
  \caption{Multi‐label performance (ROC‐AUC and F1) regarding the number of MoE experts on MIMIC‐CXR and CheXpert.}
  \label{fig:auc_f1_experts}
\end{figure}

Figure~\ref{fig:auc_f1_experts} shows how ROC‐AUC and F1‐score on MIMIC‐CXR and CheXpert vary as we increase the number of MoE experts from 2 to 16. On MIMIC‐CXR, AUC grows from 0.74 (2 experts) to 0.76 (8 experts) and remains at 0.76 for 16 experts, while F1 improves steadily from 0.78 to 0.83. CheXpert exhibits a similar pattern: AUC rises from 0.81 to 0.83 by 8 experts and then plateaus, whereas F1 climbs from 0.84 to 0.87 across the full expert range. These trends indicate that eight experts suffice to capture most of the structured‐reasoning benefits for AUC, but additional experts up to 16 continue to yield marginal F1 gains, suggesting that a larger expert bank can further refine positive‐prediction balance in multi‐label settings.

\subsection{Limitations}

\subsubsection{Domain Generalization}
% While MoE-SGT demonstrates strong performance on medical imaging (MIMIC-CXR, CheXpert) and natural image benchmarks (CUB-200, CIFAR, ImageNet), its ability to generalize across domains remains untested. Differences in concept definitions, graph structure design, and optimal hyperparameters (e.g., number of experts) may require substantial re‐tuning for applications in remote sensing, industrial inspection, or multilingual settings. Prior work has shown that neuron‐level interpretations often fail to transfer across domains \cite{sajjad2022neuron}, suggesting that our concept‐based reasoning may similarly demand careful adaptation. Future work should evaluate the robustness and transferability of our framework under varying domain shifts.
While MoE-SGT excels on medical (MIMIC-CXR, CheXpert) and natural image (CUB-200, CIFAR, ImageNet) benchmarks, its cross-domain generalization is untested. Variations in concept definitions, graph designs, and key hyperparameters (e.g., expert count) may demand substantial re-tuning for domains like remote sensing, industrial inspection, or multilingual data. Since neuron‐level interpretations often fail to transfer across domains \cite{sajjad2022neuron}, our concept-based reasoning likely requires similar adaptation. Evaluating robustness under diverse domain shifts is an important area for future work.

\subsubsection{Real-Time Dynamic Concept Generation}
Although we leverage an LLM to construct our initial concept pool, MoE-SGT currently does not support online concept expansion or adaptation at inference time. When novel concepts, pathologies, or object categories appear, the model cannot automatically incorporate them without offline retraining or manual intervention. Enabling real-time concept discovery and integration remains an important direction for making the bottleneck truly dynamic and extensible.  

\section{Conclusion}
This paper introduces MoE-SGT, a unified multimodal classification framework combining three components: a Concept Bottleneck Model (CBM) for explicit semantic predictions, a Graph Transformer over answer–concept and answer–question graphs to capture structured interactions, and a dynamic Mixture-of-Experts (MoE) module that routes inputs to specialized experts based on complexity. The interpretable concept layer ensures explainability, while graph-based reasoning models high-order cross-modal dependencies. The MoE mechanism efficiently scales capacity—assigning simple cases to smaller experts and complex ones to larger experts—yielding significant accuracy gains without sacrificing end-to-end interpretability.

\appendix

\bigskip

\bibliography{aaai2026}

%\input{ReproducibilityChecklist.tex}

\end{document}